\documentclass[conference]{IEEEtran}
\IEEEoverridecommandlockouts
\usepackage{cite}
\usepackage{amsmath,amssymb,amsfonts}
\usepackage{graphicx}
\usepackage{textcomp}
\usepackage{xcolor}
\usepackage{amsmath}
\usepackage{adjustbox}
\usepackage{algorithm}
\usepackage{algpseudocode}
\usepackage{enumitem}

\def\BibTeX{{\rm B\kern-.05em{\sc i\kern-.025em b}\kern-.08em
    T\kern-.1667em\lower.7ex\hbox{E}\kern-.125emX}}
\begin{document}

\title{Financial Fraud Detection using Jump-Attentive Graph Neural Networks}

\author{\IEEEauthorblockN{Prashank Kadam}
\IEEEauthorblockA{\textit{Vesta Corporation} \\
Atlanta, US \\
prashank.kadam@vesta.io}
}

\maketitle

\begin{abstract}
As the availability of financial services online continues to grow, the incidence of fraud has surged correspondingly. Fraudsters continually seek new and innovative ways to circumvent the detection algorithms in place. Traditionally, fraud detection relied on rule-based methods, where rules were manually created based on transaction data features. However, these techniques soon became ineffective due to their reliance on manual rule creation and their inability to detect complex data patterns. Today, a significant portion of the financial services sector employs various machine learning algorithms, such as XGBoost, Random Forest, and neural networks, to model transaction data. While these techniques have proven more efficient than rule-based methods, they still fail to capture interactions between different transactions and their interrelationships. Recently, graph-based techniques have been adopted for financial fraud detection, leveraging graph topology to aggregate neighborhood information of transaction data using Graph Neural Networks (GNNs). Despite showing improvements over previous methods, these techniques still struggle to keep pace with the evolving camouflaging tactics of fraudsters and suffer from information loss due to over-smoothing. In this paper, we propose a novel algorithm that employs an efficient neighborhood sampling method, effective for camouflage detection and preserving crucial feature information from non-similar nodes. Additionally, we introduce a novel GNN architecture that utilizes attention mechanisms and preserves holistic neighborhood information to prevent information loss. We test our algorithm on financial data to show that our method outperforms other state-of-the-art graph algorithms.
\end{abstract}


\section{Introduction}
The rise of online payment services has significantly benefited users and financial institutions but also led to a surge in financial fraud. In 2023, e-commerce fraud alone amounted to \$38 billion and is projected to exceed \$362 billion between 2023 and 2028 \cite{juniper}. Detecting various types of financial fraud, such as identity theft and credit card fraud, is crucial to mitigate these losses.

Traditional fraud detection methods relied on rule-based systems \cite{rule_based}, where manual rules flagged suspicious transactions. Initially effective, these methods became inadequate as fraudsters adapted and patterns evolved. Machine learning models like logistic regression, XGBoost, and neural networks \cite{log_reg1}, \cite{ann1}, \cite{svm1}, \cite{xg1} improved fraud detection by statistically analyzing transaction data but failed to capture interactions between transactions and struggled with evolving fraud patterns.

Graph representation learning revolutionized fraud detection by representing transaction data as graphs and using embeddings to train supervised models. Graph Neural Networks (GNNs) \cite{gnn}, \cite{gcn}, \cite{gsage} advanced the field by leveraging graph topology and node properties. Despite their success, GNNs face challenges like over-smoothing and inability to adapt to rapidly changing fraud patterns.

Graph Attention Networks (GATs) \cite{gat} improved upon this by using attention mechanisms to focus on relevant parts of the graph. However, they still struggle with over-smoothing, causing important features to be lost, especially in complex networks with family-based clusters where legitimate and fraudulent behaviors intertwine. This may lead to missed detections of camouflaging fraudsters who embed themselves within legitimate clusters.

In this paper, we propose a novel approach to address these challenges. Our method introduces a new sampling strategy and a GNN architecture called Jump-Attentive Graph Neural Network (JA-GNN). This architecture captures crucial neighborhood information and preserves key features from non-similar nodes, enhancing the system’s accuracy and responsiveness to new and evolving fraud tactics.

The key contributions of this paper are:

\begin{itemize}[leftmargin=*, topsep=0pt, partopsep=0pt, parsep=0pt, itemsep=0pt]
\item We introduce a novel sampling strategy that efficiently gathers neighborhood information and identifies camouflaged fraudsters, while also including non-similar nodes to prevent information loss.
\item We present JA-GNN, a novel GNN architecture that preserves crucial feature information from distant nodes in the final embedding. This is essential in financial fraud detection for distinguishing between legitimate and fraudulent transactions. JA-GNN incorporates residual connections from earlier layers to the output, creating a combined embedding that optimally represents the customer.
\item Our results show that our method outperforms state-of-the-art graph algorithms in detecting financial fraud.
\end{itemize}

\section{Related Work and Background}

\subsection{Financial Fraud Detection}

Financial fraud detection has long been a central research topic for financial institutions and web-based businesses. Initially, fraud detection was manual, but the rise of the Internet and increased transaction volumes made manual assessments infeasible, leading to rule-based detection methods \cite{ann1}. These methods, while easy to implement, were rudimentary, couldn't adapt to changing fraud patterns, and required expert knowledge.

To address these limitations, machine learning algorithms were applied to financial fraud detection. Classifiers like Logistic Regression \cite{log_reg1}, \cite{log_reg2}, and various configurations of Support Vector Machines \cite{svm1}, \cite{svm2}, \cite{svm3} were used. Jeragh and Al Sulaimi combined Autoencoders and One-class SVMs for credit card fraud detection \cite{svm4}. KNN-clustering, although struggling with class imbalance, was also used \cite{knn1}. Malini and Pushpa improved KNN performance by combining it with an outlier detection model \cite{knn2}. Bayesian methods like Naive Bayes \cite{bayes1}, \cite{bayes2}, and ensembles like Random Forest and XGBoost \cite{rf1}, \cite{xg1} were utilized. Hajek et al. used XGBoost with class-balancing adjustments and unsupervised outlier detection for mobile payment fraud \cite{xg1}. Artificial Neural Networks (ANN) showed significant improvements over rule-based methods but required extensive feature engineering and couldn't adapt to evolving fraud patterns \cite{ann1}.

Graph representation learning significantly advanced fraud detection. Node embeddings were initially calculated using techniques like DeepWalk \cite{deepw} and node2vec \cite{n2v}. Later methods like LINE \cite{line} and SDNE \cite{sdne} used first and second-order structural information. These embeddings were used for supervised node classification tasks but had limitations. Graph Convolutional Networks (GCNs) \cite{gcn} addressed these issues by considering both graph topology and node properties. Advanced GNN configurations like Graph Attention Networks (GAT) \cite{gat} soon followed. Wang et al. \cite{semignn} proposed a semi-supervised graph attention mechanism using multi-view data for fraud detection. Pick and Choose GNN \cite{pnc} and SCN GNN \cite{scngnn} improved fraud detection efficiency using sampling strategies and a recursive Reinforcement Learning framework. Care-GNN \cite{caregnn} tackled camouflaged fraudsters with a neighborhood sampling technique and an RL strategy. Recently, Split-GNN \cite{splitgnn} was used to detect the spectral distribution of heterophily degrees for fraud detection. GTAN \cite{gtan} demonstrated excellent results using a Gated Temporal Attention mechanism.

Despite these advancements, challenges remain. These methods struggle to adapt to dynamic fraud patterns and risk losing crucial feature information during similarity-based sampling. As fraudsters' techniques improve, these models must evolve to remain effective.

\subsection{Graph Neural Network}

Graph Neural Networks (GNNs) are a class of deep learning models tailored for graph-structured data. They operate by leveraging the graph's topology and node features to learn a vector representation for each node - capable of capturing both local and global structural information. The fundamental operation in GNNs is the neighborhood aggregation or message-passing mechanism, mathematically described as:

\begin{equation}
    h^{(n+1)}_v = \sigma\left( W^{(n)} \cdot \text{AGGREGATE}\left( \{h^{(n)}_u : u \in \mathcal{N}(v)\} \right) \right)
\label{eq:gnn}
\end{equation}

where $h^{(n)}_v$ is the feature representation of node $v$ at layer $n$, $\mathcal{N}(v)$ denotes the set of neighbors of node $v$, $W^{(n)}$ is a learnable weight matrix at layer $n$, $\sigma$ is a non-linear activation function, and $\text{AGGREGATE}$ is an aggregation function, such as sum, mean, or max, that combines the features of neighboring nodes. The initial feature representation $h^{(0)}_v$ is typically the node's raw features. By stacking multiple layers, GNNs can capture higher-order interactions between nodes, allowing for deep representation learning of graph-structured data.

\section{Model}

\subsection{Definition}

\subsubsection{Definition 1 - Transaction graph}

A transaction graph is defined as a weighted graph $\mathcal{G} = \left\{ (\mathcal{V}, \mathcal{X}, \{ \mathcal{E}_r \}_{r=1}^R), \mathcal{Y}, \mathcal{A}, \mathcal{N} \right\}$. Here, $\mathcal{V}$ is a set of all nodes in the graph represented by $\{ v_1, v_2, v_3, \dots, v_n \}$. $\mathcal{X}$ is a set of feature vectors of all the nodes given by $\{ x_1, x_2, x_3, \dots, x_n \}$ such that each vector in the set $\mathcal{X}$ has the same dimension $d$. $\mathcal{E}_r$ denotes all the edges in the graph under relationship $r$, where $r \in \{ 1, 2, \dots, R \}$. $\mathcal{Y}$ denotes a set of all the labels in the graph. Each node $v_i \in \mathcal{V}$ is associated with a label from $\mathcal{Y}$. Let $\mathcal{A} \subseteq \mathcal{V}$ represent the set of nodes with anchor labels, and let $\mathcal{N} \subseteq \mathcal{V}$ represent the set of nodes with neighbor labels, such that $\mathcal{A} \cap \mathcal{N} = \emptyset$ and $\mathcal{V} = \mathcal{A} \cup \mathcal{N}$. For every edge $(u, v) \in \mathcal{E}$, the relationship between nodes is such that $(u \in \mathcal{A}) \land (v \in \mathcal{N}) \lor (u \in \mathcal{N}) \land (v \in \mathcal{A})$. Hence, the transaction graph $\mathcal{G}$ is also a bipartite graph.

\subsubsection{Definition 2 - Weighted Mutual Neighbor Coefficient (WMNC)}

Let $\mathcal{G} = (\mathcal{V}, \mathcal{E}, w)$ be a weighted graph where $\mathcal{V}$ is the set of vertices, $\mathcal{E}$ is the set of edges and $ w: \mathcal{E} \rightarrow \mathbb{R}^+ $ is a function assigning a positive real weight to each edge in $\mathcal{E}$. The Weighted Mutual Neighbor Coefficient $\mathcal{S}_{v_1, v_2}$ between two nodes $ v_1, v_2 \in \mathcal{V} \land (v_1, v_2) \notin \mathcal{E}$ having $\mathcal{U}$ common nodes between them can then be defined as:  

\begin{equation}
\mathcal{S}_{v_1, v_2} = \frac{1}{|\mathcal{U}|} \sum_{u \in \mathcal{U}} \frac{w(v_1, u) + w(v_2, u)}{2 \times \text{min}\{\deg(v_1), \deg(v_2)\}}
\label{eq:neighbor}
\end{equation}

where $w(v_1, u)$ and $w(v_2, u)$ are the weights of the edges from $v_1$ to a common neighbor $u$ and from $v_2$ to $u$, respectively, $\deg(v_1)$ and $\deg(v_2)$ are the degrees of nodes $v_1$ and $v_2$ respectively, $\text{min}()$ represents a function that returns the minima of all the input values and $|\mathcal{U}|$ is the number of common neighbors between $v_1$ and $v_2$.

\subsection{Mutual Neighbor based sampling}

As fraud detection algorithms advance, fraudsters have adopted novel techniques to evade detection, notably camouflaging. This involves forming complex patterns to blend with non-fraudulent nodes, making their features closely resemble those of safe transactions. Identifying such camouflaged nodes is crucial. It requires detecting nodes in the neighborhood highly similar to the target, as fraudulent nodes may influence their neighbors to exhibit suspicious behavior or share similar fraudulent patterns. We propose a novel sampling strategy using the Mahalanobis distance \cite{mahalanobis} and the Weighted Mutual Neighbor Coefficient (Definition 2) to filter the $k_{top}$ nodes for prediction.

The goal is to predict the fraudulent anchors (usually the customer nodes). Thus, to sample anchors that are highly similar to the target anchor we have to go through the neighbor nodes of the target jumping two hops to the next anchor.  Each neighbor may have a different label and hence a different relationship between the anchor and the neighbors. It is essential to handle each of these relationships separately to preserve the relationship-specific information in the graph. The transaction graph is bipartite (\textit{Definition 1}) where an anchor $a_1$ is connected to another anchor $a_2$ via neighbors $n_i | i \in [1,..,N]$, and each neighbor could have a different label. 

We start with calculating the covariance matrix for each relationship in the transaction graph. This is given by:

\begin{equation}
     \sigma_{ij}^r = \frac{1}{|\mathcal{V}|_r-1} \sum_{k=1}^{|\mathcal{V}|_r} (x_{ki} - \mu_i)(x_{kj} - \mu_j) 
\label{eq:neighbor2}
\end{equation}
   
   where \(x_{ki}\) and \(x_{kj}\) are the values of dimensions \(i\) and \(j\) for the \(k\)-th embedding, and \(\mu_i\) and \(\mu_j\) are the means of dimensions \(i\) and \(j\), respectively. \(|\mathcal{V}|_r\) is the number of nodes in the graph having relationship \(r\) between them. Perform this calculation for each pair of dimensions to form the covariance matrix \(\mathbf{C}_r\). So now we have \(R\) covariance matrices for all relationships. Using this we calculate the Mahanabolis distance between all the neighbors and their adjacent edges. The Mahalanobis distance \( D_M^r \) of an anchor \( a_1 \) from a neighbor \( n_1 \) over a relationship \(r\) and covariance matrix \( \mathbf{C}_r \) is defined as:

\begin{equation}
D_M^r(a_1,n_1) = \sqrt{(a_1 - n_1)^T \mathbf{C_r}^{-1} (a_1 - n_1)}
\label{eq:neighbor3}
\end{equation}

Further we define the Edge score between the two nodes \(a_1\) and \(n_1\) as:

\begin{equation}
  E(a_1, n_1) = 
   \begin{cases} 
      1 & \text{if } D_M^r(a_1, n_1) = 0 \\
      \frac{1}{D_M^r(a_1, n_1)} & \text{if } D_M^r(a_1, n_1) > 0 
   \end{cases}
\label{eq:neighbor4}
\end{equation}

\begin{figure}[htbp]
\centering
\includegraphics[width=\linewidth]{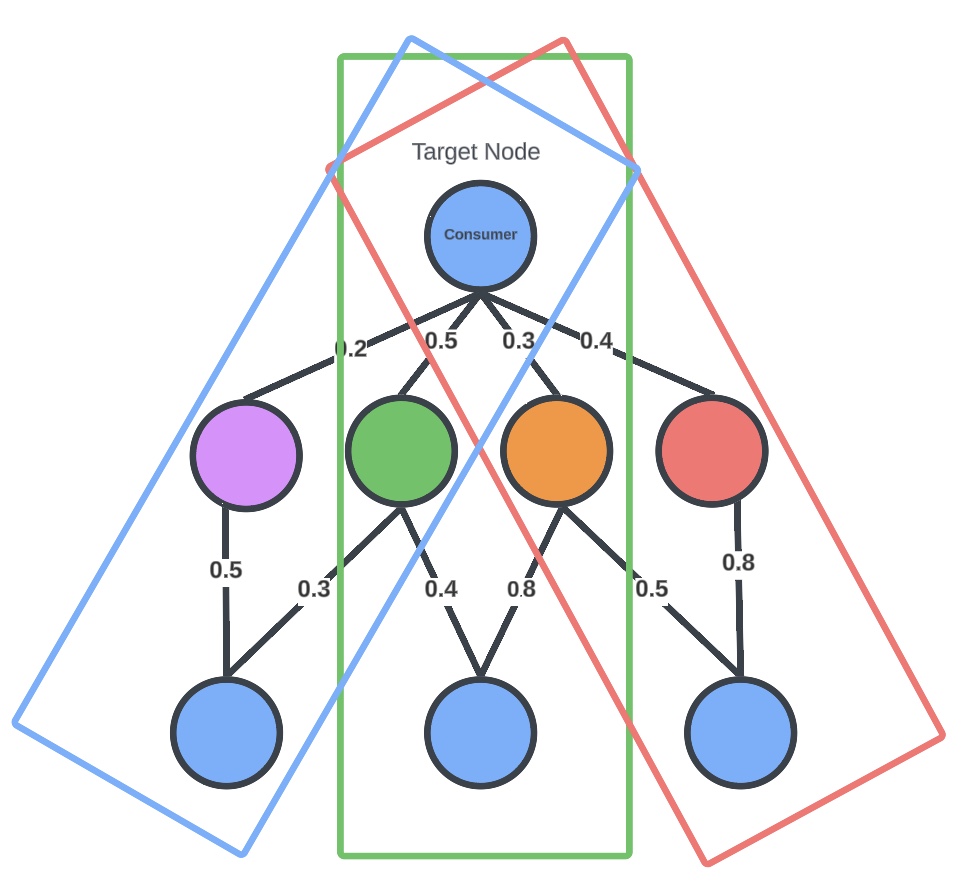}
\caption{WMNC - Each box (red, green and blue) represent the components of WMNC calculations between two anchors. }
\label{fig:mns}
\end{figure}

All the Edge scores are stored on their respective edges. Once edge scores for all the edges between $a_1$, $a_2$, and $n_i | i \in [1,..,N]$ are calculated and stored on their respective edges, we use WMNC from Definition 2 to calculate the final similarity score between $a_1$ and $a_2$ given by $S(a_1, a_2)$. Figure \ref{fig:mns} describes how WMNC is calculated. Once we have the WMNCs for all the anchor pairs, we sample the nodes from the neighborhood of the target node to keep only those nodes that are highly similar to the target. We select the $k_{top}$ nodes using the Silhouette Score, which is a metric designed to measure how similar a node is to its own cluster compared to other clusters.

Considering anchor $a_i$ as the target node, we define  k  as the set of all nodes at a two-hop distance from $a_i$, the Silhouette Score for a set of k nodes can be given by:

\begin{equation}
\text{Silhouette}(k) = \frac{1}{k} \sum_{j=1}^{k} \frac{b(a_j) - a(a_j)}{\max(a(a_j), b(a_j))}
\label{eq:neighbor5}
\end{equation}

where $a(a_j)$ is the average distance between node $a_j$ and all other nodes in the same cluster, and $b(a_j)$ is the minimum average distance from node $a_j$ to nodes in a different cluster.

To find the optimal k, we compute the Silhouette Score for different values of k and select the one that maximizes this score. The optimal $k_{top}$ can be given by:

\begin{equation}
k_{top} = \arg\max_{k} \text{Silhouette}(k)
\label{eq:neighbor6}
\end{equation}

This equation represents the task of selecting the subset \( k \) of \((N^{2*}(a_1) \cup a_1)\), where \(N^{2*}(a_1)\) are the nodes present at two-hop neighborhood of node \(a_1\), that maximizes the Silhouette Score, indicating the best clustering configuration for node \(a_i\) and its neighborhood.

The $k_{top}$ similar neighbors would have features very similar to node $a_i$ which will help in camouflage detection. There might exist a feature on one of the non-similar nodes that is a strong identifier of fraud. This would not be a part of the $k_{top}$ nodes and thus we would miss out on this crucial information. Hence we introduce random sampling from the remaining nodes. Thus, the sampled sub-graph can be given by:

\begin{equation}
N_{random} = \epsilon . N^{2*}(a_i) \setminus k_{top} 
\label{eq:neighbor7}
\end{equation}

Here $a_i$ is the node whose neighborhood is being sampled, and $N^{2*}(a_i)$ are all the nodes at a two-hop distance from $a_i$. $\epsilon$ is the randomness coefficient which is a constant that allows us to sample $N_{random}$ nodes from the neighborhood.  Note that the final sampled sub-graph contains the $k_{top}$, $N_{random}$ and $a_i$ anchor nodes along with their one-hop neighbors which are all the neighbor nodes (nodes with labels apart from the anchor label) represented by $N^1(k_{top}, N_{random}, a_i)$ in the following equation.

\begin{equation}
\centering
\begin{gathered}
    N_{sampled}(a_i) = k_{top} \cup N_{random} \cup a_i \cup N^1(k_{top}, N_{random}, a_i)
\label{eq:neighbor8}
\end{gathered}
\end{equation}

\subsection{Jump-Attentive Graph Neural Network (JA-GNN)}

This section defines the Graph Neural Network (GNN) architecture designed to leverage neighborhood information to distinguish between fraudulent and non-fraudulent transactions. Graph Attention Networks (GATs) \cite{gat} use an attention mechanism to focus on relevant graph parts, aiding fraud detection. However, GATs suffer from over-smoothing, where node features become overly homogenized, risking the approval of obvious fraud instances by losing crucial information.

We propose JA-GNN to address this issue. Attention-based mechanisms prioritize similar neighbors, aiding in detecting camouflaged fraudsters but often ignoring non-similar nodes that might indicate fraud. For instance, in credit fraud modeling, a chargeback signal at a 2-3 hop distance is critical but might be missed if it has a low attention coefficient.

JA-GNN creates a jumping mechanism between \(d\) attention layers and the output, preserving crucial features from previous layers in the final embedding. This prevents the loss of important information, such as a distant chargeback signal, due to over-smoothing. It ensures all similarity-based node information is conserved in the final embedding by providing attention to only the top \(k\) nodes. JA-GNN consists of two main parts: an aggregation mechanism to focus on important neighborhood information and a jumping mechanism to preserve critical feature information from \(d\) previous layers.

\subsubsection{\textbf{Attention Mechanism}}
In this part, we focus on maximizing the similarity between two nodes and thereby amplifying the properties of highly similar nodes. Consider an \(\mathsf{N}\)-layered GNN, where the output embedding of each layer for a node \(v\) can be given by \(h_v^n\) for \(n \in [1,..., \mathsf{N}]\). Layer \(\mathsf{N}\) produces the final embedding. In a JA-GNN, each layer has an attention mechanism where there exists an attention vector \(a^T_{n,r}\) for \(n \in [1,...,\mathsf{N}]\) and $r \in [1,.., R]$ where $\mathsf{N}$ and $R$ represent the total number of layers and relationships respectively. This attention vector would maximize the similarity between the given node and its neighbors over a given relationship. We calculate the attention coefficient between the node and each of its neighbors over a given relationship at layer $k$ with the following formula:

\begin{equation}
    \alpha^r_{vu} = softmax_u (LeakyReLU (a_{n,r}^T [h_{v}^r || h_{u}^r]))
\label{eq:jagnn1}
\end{equation}

Here \(\alpha^r_{vu}\) is the attention coefficient over relationship $r$, \(h_{v}^r\) and \(h_{u}^r\) are the embedding of the given node and its corresponding neighbor over relationship $r$ respectively, \(||\) represents a concatenation operation and finally \(a^T\) represents the attention vector for the given layer. Higher the similarity between between \(h_{v}^r\) and \(h_{u}^r\), higher would be the value of \(\alpha^r_{vu}\). The final embedding of the aggregation mechanism \( h^{agg}_v \)  can be given by:


\begin{equation}
\scriptsize
\begin{split}
\mathbf{h}_v^{agg} = \sigma \Bigg( \sum_{r=1}^R \Bigg( \sum_{u \in (\mathcal{N}(v,r) \cap k_{\text{top}}^{(n-1)}(v,r)) \cup \{v\}} \alpha_{vu}^{(n,r)} \mathbf{W}^{(n,r)} \mathbf{h}_u^{(n-1,r)} \Bigg) \Bigg), \\
\text{for } n = [1,N]
\end{split}
\label{eq:jagnn2}
\normalsize
\end{equation}

where:\(\mathbf{W}^{(n,r)}\) is the weight matrix for layer, \(\mathbf{h}_u^{(n-1,r)}\) is the embedding of node u from the previous layer, \(\alpha_{vu}^{(n,r)}\) is the attention coefficient between nodes $v$ and $u$ at layer $n$; all these are over a relationship $r$.\(\sigma\) is a non-linear activation function, like ReLU. $\mathcal{N}(v,r) \cap k_{\text{top}}^{(n-1)}(v,r)) \cup \{v\}$ represents the neighbors of node $v$ from the previous layer that were a part of the $k_{top}$ nodes sampled in Equation \ref{eq:neighbor6} including node $v$. We use multi-headed attention and average out the final embedding values after each layer \(n\):

\begin{equation}
\scriptsize
\begin{split}
\mathbf{h}_{v, n}^{agg} = \sigma \left( \frac{1}{M} \sum_{m=1}^{M} \sum_{r=1}^R \sum_{u \in (\mathcal{N}(v,r) \cap k_{\text{top}}^{(n-1)}(v,r)) \cup \{v\}} \alpha_{vu}^{m,r} \mathbf{W}^{(n,r)} \mathbf{h}_u^{(n-1,r)} \right)
\end{split}
\label{eq:jagnn3}
\normalsize
\end{equation}

\subsubsection{\textbf{Jumping Mechanism}}
In this part, we develop a way to conserve the complete neighborhood information in the final embedding. The goal is to ensure that even if critical markers on dissimilar nodes might be obscured due to over-smoothing from attention-based mechanisms, their information is still preserved in the final embedding. We create jump connections between layers \(\mathsf{N} - d\) to \(\mathsf{N}\) where \(d\) is a parameter that represents the depth up to which we want to create the jump connections. We learn \(d\) during training by using the Progressive Deep Learning \cite{progdl} approach. We start the process with the smallest possible value of \(d\) (1 in this case) and progressively increase the value to the maximum possible value (\(\mathsf{N}\) in this case) to find the value for which the network performs optimally, we then use that value at inference. Additionally, we learned a distribution over $d$ based on the input using a single-layer MLP to estimate $d$, thus dynamically adjusting the network architecture. However, this approach did not perform well as it increased inference time, and given that GNNs are inherently shallow, there was no performance improvement. 

At a layer \(n\), consider a node \(v\) having neighbors \(N(n,r)\) over a relationship $r$ and \(n \in [\mathsf{N}-d, ..., \mathsf{N}]\). We sample $N_{random}(n,r)$ nodes in \(N(n,r)\) as per Equation $\ref{eq:neighbor7}$ to fetch the embedding of the nodes which are most dissimilar to $v$. The jump-based embedding for this layer can then be given by \(h^{jump}_{v,n} = \sum_{r=1}^R \sum_{u \in N_{random}(v,r)}  \frac{h_u^{n-1,r}}{|N_{random}(v,r)|}\) which is simply the average of all the sampled nodes over all the relationships. The final output jumping embedding would therefore be:

\begin{equation}
\begin{split}
h^{jump}_v = AGG \left( \sum_{r=1}^R \sum_{u \in N_{random}(v,r)} \frac{h_u^{n-1,r}}{|N_{random}(v,r)|} \right); \\
n \in [\mathsf{N}-d, \dots, \mathsf{N}]
\end{split}
\label{eq:jagnn4}
\end{equation}
     
\(AGG\) represents some aggregation schema like concatenation, max pooling, averaging, etc.

\subsubsection{\textbf{Combined Embedding}}

\begin{figure*}[htbp]
\centering
\includegraphics[width=\linewidth]{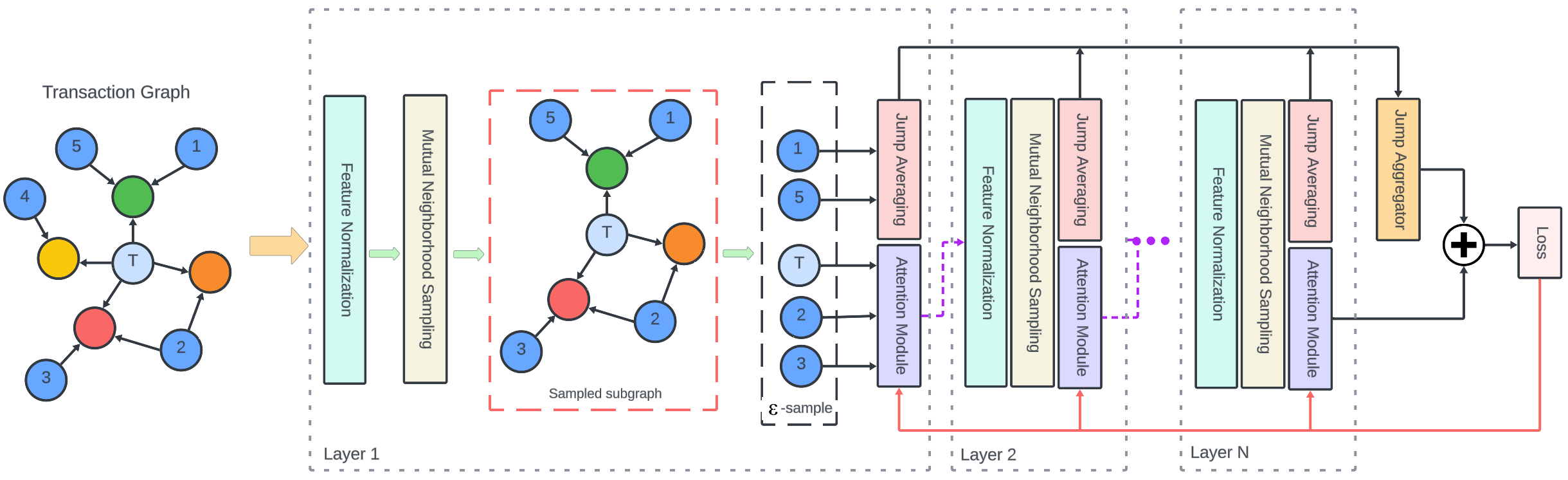}
\caption{JA-GNN architecture}
\label{fig:jagnn}
\end{figure*}

The combined embedding can thus be given by a concatenation of \(h^{agg}_v\) and \(h^{jump}_v\), Figure \ref{fig:jagnn} describes the complete JA-GNN architecture. Details of the JA-GNN algorithm are given in algorithm \ref{algo:jagnn}. 
\begin{equation}
   h_v = h^{agg}_v || h^{jump}_v
\label{eq:jagnn5}
\end{equation}

\subsubsection{\textbf{Loss Function}}

We use binary cross-entropy loss for training our JA-GNN can be given as follows:

\begin{equation}
    L_{\text{classification}} = -\frac{1}{|\mathcal{V}|} \sum_{i=1}^{|\mathcal{V}|} \left[ y_i \log(\hat{y}_i) + (1 - y_i) \log(1 - \hat{y}_i) \right]
\label{eq:loss1}
\end{equation}

Where $|\mathcal{V}|$ is the number of nodes in the graph. $y_{i}$ is the true label for node $i$, where $y_i \in \{0, 1\}$. $\hat{y}_{i}$ is the predicted probability for node $i$ to belong to class 1. $d$ is the depth of the jumping mechanism starting from the output layer. $\lambda$ is the regularization coefficient. The regularization term for penalizing the number of residual connections is:
\begin{equation}
    L_{\text{regularization}} = \lambda d
\label{eq:loss2}
\end{equation}

The idea of introducing this regularization term is that for higher values of \(d\), graph representations further away from the current node will be concatenated with the node representations of the current node, this may lead to an over-representation of distant neighbors which may lead to loss of information in the attention-based representation of the nodes. Hence, we penalize loss for larger values of \(d\). We have kept the penalty linear as it showed the best experimental results between log, linear, and exponential penalties. The final combined loss function is the sum of the classification loss and the regularization term:
  \[  L_{\text{final}} = L_{\text{classification}} + L_{\text{regularization}} \]

\begin{equation}
   L_{\text{final}} = -\frac{1}{|\mathcal{V}|} \sum_{i=1}^{|\mathcal{V}|} \left[ y_i \log(\hat{y}_i) + (1 - y_i) \log(1 - \hat{y}_i) \right] + \lambda d
\label{eq:loss3}
\end{equation}

\begin{algorithm}[h]
\caption{Training Algorithm for JA-GNN}
\label{algo:jagnn}
\begin{algorithmic}[1]
\Require The transaction graph: $\mathcal{G} = \{ (\mathcal{V}, \mathcal{X}, \{(\mathcal{E}_r)\}_{r=1}^R), \mathcal{Y}, \mathcal{A}, \mathcal{N} \}$. The number of Epochs, Batches, and  Layers = $E, B, \mathsf{N}$. The sampling constant, randomness constant = $k, \epsilon$. The jump depth $d$ and total depth $D$. The regularizer weight $\lambda$.
\Ensure The model parameters $\Theta$ and the attention parameter $a$
\State Randomly initialize the model parameters $\Theta$ and the attention parameter $a$.
\For{$e = 1, ...., E$}
    \For{$b = 1, ...., B$}
        \For{$l = 1, ...., \mathsf{N}$}
            \State Fetch the $D$ hop neighborhood of each transaction and create a sub-graph
          \For{$r = 1, ...., R$}
            \State Get $E(v, v')$ using using Eq. \ref{eq:neighbor2}, \ref{eq:neighbor3} and \ref{eq:neighbor4}
            \State Get $S(v, v')$ is Eq. \ref{eq:neighbor}
            \State Get $N_{sampled}$ using Eq. \ref{eq:neighbor6}, \ref{eq:neighbor7} and \ref{eq:neighbor8}
          \EndFor
          \State Get $h^{agg}_{v}$ using Eq. \ref{eq:jagnn3}
        \EndFor
        \State  Get $h^{jump}_v$ using Eq. \ref{eq:jagnn5}
        \State  Get $h_v$ using Eq. \ref{eq:jagnn5}
        \State Get $L_{final}$ using Eq. \ref{eq:loss3} 
        \State Update model parameters and attention parameters $\Theta$ and $a$ using backpropagation
    \EndFor
\EndFor
\State \Return $\Theta, a$
\end{algorithmic}
\end{algorithm}

\section{Experiments}

In this section, we will be performing experiments based on real-world data to evaluate the performance of our models.

\subsection{Dataset}

\subsubsection{PFFD} - Our dataset, called the Proprietary Financial Fraud Dataset (PFFD), is sourced from our organization's e-commerce clients. Each row in the PFFD represents a customer order, containing hashed information on Name, Address, Email, Phone, Device, Payment details, etc. ensuring customer privacy.

The transaction graph is customer-centric, where each customer node is the anchor, connected to attribute nodes. It is a bipartite graph following the schema: \textit{Customer -$>$ [Attributes (Name, Address, Phone, Email, Device, Payment)] $<$- Customer}. Continuous properties like transaction amounts are updated by aggregating values on respective nodes, eliminating the need for feature engineering.

Our target label, Fraud Score (fs), ranges from 0 to 100, indicating the likelihood of fraud. The dataset includes approximately 1.25 million transactions, with 13,454 fraudulent cases, making up 1.07\% of the data. The graph's average node degree is 14.71, with around 29,000 hypernodes (degree $>$ 100). During training, hypernodes are undersampled to 100 immediate neighbors before Mutual Neighborhood-based sampling identifies the $k_{top}$ neighbors.

\subsubsection{Yelp Dataset} - Although our method is primarily developed for financial data, we have tested it on other popular open-source datasets, such as the Yelp-Fraud dataset \cite{caregnn}. This dataset includes all reviews posted by users on Yelp, with the task of identifying fraudulent reviews. The relationships modeled in the graph, such as R-U-R, R-S-R, and R-T-R, fit seamlessly with our definition of a bipartite transaction graph (\textit{Definition 1}). We applied the same approach used for modeling our proprietary graph to train our model on the Yelp-Fraud dataset.

\subsection{Training} 
Training JA-GNN involves mini-batches with equal numbers of fraud and non-fraud transactions. The data is chronologically sorted and split into 60:20:20 for train, validation, and test sets to preserve changing fraud patterns. Each mini-batch creates a sub-graph of nodes and their neighbors, with a feature normalization layer added before each GNN layer to standardize dimensionality. Sampling follows the method in section 3.2, with depth determined by JA-GNN’s layers. Fraudulent transaction nodes have their fraud scores set to 100. We use the precision-recall curve to evaluate performance, selecting the optimal threshold by maximizing the F1-score. Hyperparameter tuning is performed during validation.

\subsection{Baselines} 
To assess the effectiveness of JA-GNN in mitigating the impact of camouflaged fraudsters, we compared it with various GNN baselines under a semi-supervised learning framework. From a general GNN architecture perspective, we use GCN \cite{gcn}, RGCN \cite{rgcn}, GAT \cite{gat}, and GraphSAGE \cite{sage}. For state-of-the-art comparisons we use CareGNN \cite{caregnn}, Semi-GNN \cite{semignn}, PC-GNN\cite{pnc}, and GTAN\cite{gtan} since these are some of the best-performing methods on publicly available datasets like Yelp Fraud. To analyze the performance of each of our contributions we have implemented three variants of JA-GNN ($\text{JA-GNN}_{sample}$, $\text{JA-GNN}_{solo}$, $\text{JA-GNN}$), details of each are mentioned in the ablation study section.

\subsection{Experimental Setup} 
All the transactions are loaded to the Neo4j graph database and neighborhoods are extracted for minibatches using Cypher queries. To prevent linkages based on future data, the Neo4j graph was loaded with one data set at a time. First, the training data was loaded, once model training was done, the graph was updated with validation data and finally, test data was loaded part by part. Once the data is extracted, it is converted to a format compatible with PyTorch Geometric. For GCN \cite{gcn}, RGCN \cite{rgcn}, GAT \cite{gat}, GraphSAGE \cite{sage}, PC-GNN \cite{pnc} and GTAN \cite{gtan} we use implementations given by the authors. We utilize available open-source implementations for SemiGNN \cite{semignn} and CareGNN \cite{caregnn}. Model training is done on the Databricks platform. We use a 14.3 LTS ML runtime environment with NC12 v3 instance supporting 2 NVIDIA Tesla V100 GPUs with 224 Gigabytes of RAM. 

The JA-GNN model uses the following hyperparameters after training and tuning. The embedding length is 128, the batch size we use is 1024, the number of layers in JA-GNN is 3, and the learning rate is 0.01 with the Adam optimizer. The regularization weight $\lambda$ is 0.0002. Jumping depth $d$ is 3, and we chose this value after testing will multiple values of $d$, the jump aggregation function that we use for our experiments is the mean function.

Our dataset is highly imbalanced, with only 1.07\% of the data labeled as fraud. The final task is to identify the fraudsters. Due to this, we will be using ROC-AUC and Recall as evaluation metrics for recording the performance of each of our models.

\subsection{Ablation Study} 
As a part of this paper, we have two main components - Mutual neighborhood sampling and JA-GNN. In this section, we evaluate the performance of each of these components:

\begin{itemize}[leftmargin=*, topsep=0pt, partopsep=0pt, parsep=0pt, itemsep=0pt]
    \item $\text{JA-GNN}_{sample}$ - In this architecture, only the mutual neighborhood sampling is used during training. The GNN model is kept attention-based but does not have any residual connections.
    \item $\text{JA-GNN}_{solo}$ - In this architecture, we do not use the mutual neighborhood sampling. This architecture only consists of the JA-GNN with jump connections to the output layer from layers at depth $d$. The complete neighborhood of the target node is considered for both attention and jump-based embedding.
    \item $\text{JA-GNN}$ - This is the complete architecture, which consists of all components defined in this paper. 
\end{itemize}

\subsection{Results}


\begin{table*}[htbp]
\centering
\caption{Fraud detection evaluation}
\begin{tabular}{|p{0.06\textwidth}||p{0.05\textwidth}||p{0.05\textwidth}|p{0.05\textwidth}|p{0.05\textwidth}|p{0.05\textwidth}|p{0.05\textwidth}|p{0.04\textwidth}|p{0.05\textwidth}|p{0.07\textwidth}|p{0.07\textwidth}|p{0.07\textwidth}|p{0.07\textwidth}|}
\hline
\textbf{Dataset} & \textbf{Metric} & \textbf{GAT} & \textbf{RGCN} & \textbf{Graph SAGE} & \textbf{Semi-GNN} & \textbf{CARE-GNN} & \textbf{PC-GNN} & \textbf{GTAN} & \textbf{JA-GNN \textit{sample}} & \textbf{JA-GNN \textit{solo}} & \textbf{JA-GNN}\\
\hline
\hline
PFFD & AUC & 0.784 & 0.771 & 0.762 & 0.811 & 0.793 & 0.799 & 0.817 & 0.791 & 0.832 & \textbf{0.897} \\
& Recall & 0.765 & 0.764 & 0.743 & 0.787 & 0.776 & 0.795 & 0.801 & 0.769 & 0.816 & \textbf{0.868}\\
\hline
Yelp & AUC & 0.572 & 0.553 & 0.587 & 0.577 & 0.736 & 0.801 & 0.901 & 0.689 & 0.917 & \textbf{0.951} \\
& Recall & 0.547 & 0.512 & 0.541 & 0.549 & 0.697 & 0.773 & 0.836 & 0.628 & 0.877 & \textbf{0.917}\\
\hline
\end{tabular}
\label{table1}
\end{table*}

The proposed fraud detection methods were evaluated on PFFD and Yelp datasets, comparing JA-GNN variants and other state-of-the-art models using AUC and Recall metrics (see Table \ref{table1}).

\subsubsection{PFFD} - The JA-GNN models demonstrated superior performance on the PFFD. JA-GNN achieved the highest AUC of 0.897, indicating strong discriminatory capability. This was followed by $\text{JA-GNN}_{solo}$ with an AUC of 0.832 and GTAN with an AUC of 0.817. In recall, JA-GNN scored 0.868, $\text{JA-GNN}_{solo}$ 0.816, and GTAN 0.801, showing JA-GNN's effectiveness in identifying true fraudulent cases.

\subsubsection{Yelp Dataset} - On the Yelp dataset, JA-GNN models also showed a clear advantage. JA-GNN had the highest AUC of 0.933, followed by JA-GNN \textit{solo} with 0.917 and GTAN with 0.901. JA-GNN achieved a recall of 0.906, $\text{JA-GNN}_{solo}$ 0.877, and GTAN 0.836, indicating JA-GNN's superior performance in detecting fraudulent reviews.

Thus we see the JA-GNN outperforms all the current state-of-the-art fraud detection algorithms. Even $\text{JA-GNN}_{solo}$ (without sampling) was able to perform better than GTAN which is the currently best performing algorithm in the literature. $\text{JA-GNN}$ (with sampling) was able to further outperform $\text{JA-GNN}_{solo}$ thus demonstrating the effectiveness of each component and the complete algorithm.

Table \ref{table2} shows how JA-GNN performed at different values of $d$, we see that we get the most optimal performance at $d$=2.

\begin{table}[H]
\centering
\caption{Jump Attentive GNN analysis}
\begin{tabular}{l|ccc}
\hline
\textbf{Jump depth} ($d$) & \textbf{AUC} & \textbf{Recall} \\
\hline
0 & 0.809 & 0.797 \\
1 & 0.821 & 0.805 \\ 
2 & 0.847 & 0.828 \\
3 & 0.824 & 0.810 \\ 
\hline
\end{tabular}
\label{table2}
\end{table}

\subsection{Evidence of Camouflage Detection}

\begin{figure}[h]
\centering
\includegraphics[width=250px]{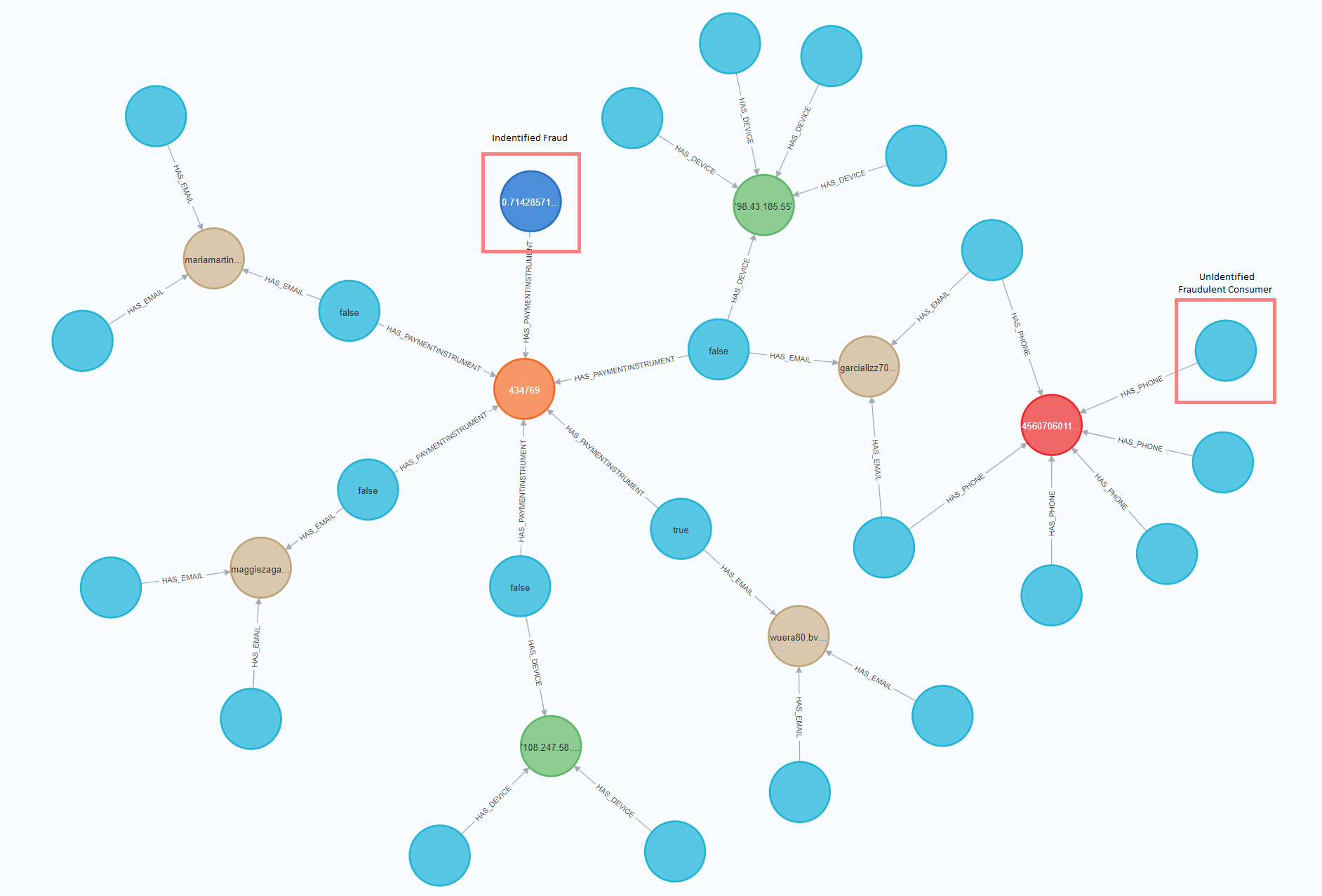}
\caption{Camouflage: The figure shows camouflaged fraudsters, with the dark blue node indicating a verified fraudster and the light blue node indicating a newly ingested, unverified fraudulent transaction.}
\label{fig:neo4j_fraud}
\end{figure}

Most state-of-the-art fraud detection algorithms assume high similarities between camouflaging fraudsters and their non-fraudulent neighbors, leading to over-smoothing and loss of critical information. In one of our production graphs (figure \ref{fig:neo4j_fraud}), a dark blue node represents a verified fraudster within a subgraph of six individuals using similar details like credit cards and phone numbers, resulting in multiple customer nodes.

The dark blue node has a “chargeback” feature set to True, indicating past fraudulent activity. A new fraudulent node attaches to the subgraph but is not yet identified as fraud. We generate embeddings for this new node using different algorithms and obtain a fraud score between 0 to 100 (see table \ref{table-fraud}). JA-GNN predicted the highest fraud score, demonstrating its effectiveness against camouflaging fraudsters, a behavior seen in about 20\% of financial fraud cases.

\begin{table}[H]
\centering
\caption{Evidence of Camouflage detection}
\begin{tabular}{|p{0.04\textwidth}||p{0.04\textwidth}|p{0.04\textwidth}|p{0.04\textwidth}|p{0.04\textwidth}|p{0.04\textwidth}|p{0.04\textwidth}|p{0.06\textwidth}|}
\hline
\textbf{Metric} & \textbf{GAT} & \textbf{Graph SAGE} & \textbf{Semi-GNN} & \textbf{CARE-GNN} &  \textbf{GTAN} & \textbf{JA-GNN}\\
\hline
Fraud Score & 36.43 & 40.21 & 44.30 & 41.64 & 53.32 & \textbf{61.16}  \\
\hline
\end{tabular}
\label{table-fraud}
\end{table}

\section{Conclusion}
In this paper, we propose a novel algorithm to improve the detection of financial fraud. We introduce a new sampling strategy that can efficiently sample camouflaged neighbors as well as introduce randomness to the sample to pick on other nodes that might be storing crucial features that could determine immediate fraud. We introduce a novel GNN architecture, JA-GNN, that creates jump connections from a fixed depth to the output layer to create a holistic neighborhood representation by making use of our sampling technique. Our results show strong performance improvements over the current state-of-the-art models. As a future scope for this research, we are working on curated curricula consisting of specific fraud patterns to dynamically adjust the number of jump connections at runtime.




\end{document}